\documentclass[runningheads]{llncs}
\usepackage[T1]{fontenc}
\usepackage{graphicx}
\usepackage{hyperref}
\usepackage{indentfirst}
\usepackage[ruled, noline, longend, linesnumbered]{algorithm2e}
\usepackage{booktabs}
\usepackage{tabularx}
\usepackage{makecell}
\usepackage{multirow}
\usepackage{tablefootnote}
\usepackage{newfloat}
\usepackage{listings}
\usepackage{amsmath}
\usepackage{verbatim}
\usepackage{graphicx}
\usepackage{adjustbox}
\usepackage{amssymb}
\usepackage{amsmath}
\usepackage{xcolor}

\begin{document}
\title{LLM4Rail: An LLM-Augmented Personalized Railway Services Platform}
%
 \author{
 Zhuo Li\inst{1}\orcidID{0000-0002-2141-788X} \and
 Xianghuai Deng\inst{1}\orcidID{0009-0006-2893-8211} \and 
 Chiwei Feng\inst{1}\orcidID{0009-0005-4999-3474} \and
 Hanmeng Li\inst{1} \and
 Shenjie Wang\inst{1} \and
 Junchao Fan\inst{1}\orcidID{0000-0003-2586-7973} \and  
 Jiaxing Shen\inst{2}\orcidID{0000-0002-0833-0288} \and
 Zhiyuan Wen\inst{3}\orcidID{0000-0001-5644-5023} \and
 Longxiang Gao\inst{4} \and
 Linchun Wu\inst{5}
 } 
 \authorrunning{Z. Li, X-H. Deng, C-W. Feng, et al.}
 \institute{
 School of Artificial Intelligence, Chongqing University of Posts and Telecommunications, Chongqing, China \\
 \email{lizhuo@cqupt.edu.cn}, \email{\{xianghuai.deng518,fengchiwei.edu\}@gmail.com}, \email{\{hanmeng.li521,shenjie.wang1107\}@gmail.com}, \email{fanjc@cqupt.edu.cn} \and
 Division of Artificial Intelligence, Lingnan University, Hong Kong, China \\
 \email{jiaxingshen@ln.edu.hk} \and
 Department of Computing and The Institute for Higher Education Research and Development, The Hong Kong Polytechnic University, Hong Kong, China \\
 \email{zyuanwen@polyu.edu.hk} \and
 Faculty of Computer Science and Technology, Qilu University of Technology, Jinan, China \\  
 \email{gaolx@sdas.org} \and
 Chengdu Youqi Zhihong Technology Co., Ltd., Chengdu, China \\
 \email{wulc505@126.com} 
 }
\maketitle              
\begin{abstract}
Large language models (LLMs) have significantly reshaped different walks of business. 
To meet the increasing demands for individualized railway service, we develop LLM4Rail\footnote{We release our code at \href{https://www.github.com/CQUPTAI/LLM4Rail}{https://www.github.com/CQUPTAI/LLM4Rail}}\,\textbf{--}\,a novel LLM-augmented personalized railway services platform.
Empowered by the LLM, LLM4Rail provides dedicated modules for ticket and weather inquiries, food \& drink recommendations, and casual conversation.
In LLM4Rail, we propose the iterative ``Experience-Retrieval Question-Thought-Action-Observation (ER-QTAO)'' reasoning framework.
It introduces two key mechanisms, Experience Storage and Retrieval (ESR) and Diagnostic Observation Generation (DOG), to enhance the LLM's ability to invoke appropriate external tools with complex arguments.
To provide personalized onboard dining services, we first construct the 
Chinese Railway Food \& Drink (CRFD) dataset, which covers a wide range of 
signature dishes categorized by city, cuisine, age group, and spiciness 
level, and then develop an LLM-based zero-shot conversational recommender system  
for railway catering.
In addition, to address the unconstrained nature of generative recommendations, we design a feature-similarity-based alignment mechanism to ensure that all recommended items are aligned with entries in the CRFD dataset.

\keywords{LLM4Rail \and Large Language Models (LLMs) \and ER-QTAO \and CRFD Dataset \and Zero-shot Conversational Recommender System.}
\end{abstract}
\section{Introduction}

Railway companies adhere to the customer-first principle. 
However, despite the rapid expansion of China's railway infrastructure over the past decade, service quality remains a critical area for improvement. 
While face-to-face customer service is available at train stations and on board, online services have yet to fully meet evolving passenger expectations
In the mobile internet era, the China Academy of Railway Sciences (CARS) launched \href{https://play.google.com/store/apps/details?id=com.chinarailway.globalticketing&hl=en&pli=1}{Railway\,12306} platform, which provides basic ticketing functionalities. 
Nevertheless, amid the recent global economic downturn, railway companies are actively seeking new revenue streams. 
Following consultations with CARS, we identified a pressing demand for integrating value-added services into the Railway 12306 App, particularly personalized food and beverage ordering services for onboard dining.
Motivated by this, we build LLM4Rail, an LLM-augmented platform designed to deliver comprehensive railway-related services\,\textbf{--}\,including train schedules, ticket information, weather updates, onboard catering recommendations, and casual conversation\,\textbf{--}\,through a seamless, user-centric interface.

The emergence of large language models (LLMs), such as ChatGPT~\cite{InstructGPT,GPT3,GPT4,gpt-4o}, Qwen~\cite{qwen,qwen3,qwenvl}, and DeepSeek~\cite{deepseek-v3,deepseek-r1}, has profoundly reshaped both industry and academia. 
However, base LLMs often underperform in domain-specific tasks due to insufficient proprietary training data. 
To overcome the limitations of general-purpose LLMs in railway applications, 
LLM4Rail follows the paradigms of ReAct~\cite{react} and 
Reflexion~\cite{reflexion}, incorporating external tools to deliver accurate 
and reliable responses.
Nonetheless, vanilla ReAct-style tool orchestration struggles to generate high-quality arguments for tools with complex parameter schemas, whereas Reflexion not only consumes a substantial portion of the context window as the number and complexity of tools grow, but also lacks a mechanism to revise outdated memories in unstable environments.
To address these issues, LLM4Rail employs an ``Experience-Retrieval Question-Thought-Action-Observation (ER-QTAO)'' framework, which introduces two key mechanisms: (i) Experience Storage and Retrieval (ESR), which summarizes tool-use experiences from historical interactions, retrieves relevant experiences to guide current reasoning, and discards obsolete experiences in a timely manner, and (ii) Diagnostic Observation Generation (DOG), which transforms external tool's raw outputs into structured diagnostic reports to guide subsequent actions and experience summarization.

Traditional task-oriented dialogue (TOD) systems~\cite{e2e_tod} follow a pipeline of natural language understanding (NLU)~\cite{attention_nlu,slot_nlu,cap_nlu,bert_nlu}, dialogue state tracking 
(DST), dialogue policy, and natural language generation (NLG)~\cite{tod_review}.
The subsequent end-to-end approach unifies these modules into a single neural model. 
However, both paradigms rely heavily on large-scale labeled dialogue data, and the training process is costly and time-consuming. 
In contrast, LLM4Rail performs dialogue understanding and intent recognition in a zero-shot manner by leveraging the inherent natural language comprehension capabilities of the base LLMs.
Upon identifying passengers' travel needs, the LLM4Rail agent invokes appropriate tools to fulfill railway-specific tasks, ranging from ticket inquiries, weather updates, food \& beverage recommendations to casual conversation.
The ticketing module performs fuzzy search to return the top-3 matching tickets, and the weather information retrieval module provides real-time weather updates.
The food \& drink component provides customized onboard dining services tailored to individual preferences, supported by the Chinese Railway Food \& Drink (CRFD) dataset\,---\,the first railway-specific takeout dataset encompassing diverse dishes across cities, cuisines, and price ranges. 
If passengers inquire about topics beyond the aforementioned scope, the agent initiates casual conversation by leveraging the intrinsic capabilities of the base LLMs.

\indent In summary, the key contributions of this work are threefold: \\
\indent $\bullet$ Develop a railway-specific LLM-based platform, delivering personalized services including weather, ticketing, and onboard food \& drink recommendations.   \\
\indent $\bullet$ Propose the Experience-Retrieval QTAO (ER-QTAO) framework, which utilizes Experience Storage and Retrieval (ESR) and Diagnostic Observation Generation (DOG) to enhance the LLM’s ability to select and invoke suitable tool APIs with complex input parameters.   \\
\indent $\bullet$ Construct the Chinese railway-oriented takeout dataset\,\textbf{--}\,Chinese Railway Food \& Drink (CRFD), and devise an LLM-based zero-shot conversational recommender with feature alignment for food \& drink recommendation.

\section{LLM4Rail Platform}

\begin{figure*}[htbp]
    \hspace{-2cm}
    \includegraphics[width=16cm]{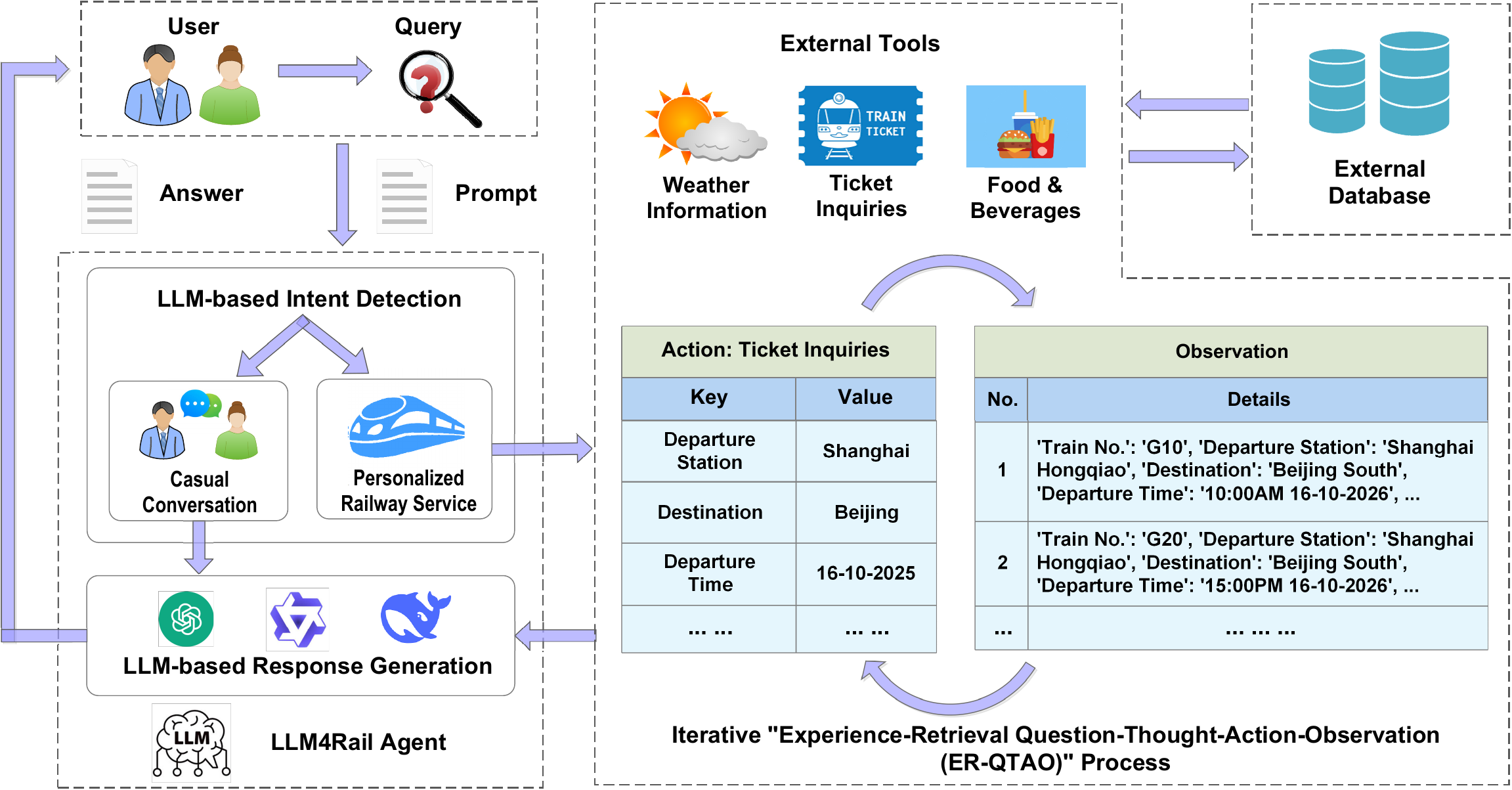}
    \caption{The overall architecture of the LLM4Rail platform.} 
    \vspace{-0.3cm}
    \label{fig:illustration_of_workflow}
\end{figure*}

Fig.~\ref{fig:illustration_of_workflow} illustrates the overall architecture of the LLM4Rail platform.
It provides various personalized railway services (including ticketing, weather information retrieval, and food \& drink recommendations), as well as casual conversation.
LLM4Rail also introduces a novel iterative reasoning framework, termed ``Experience-Retrieval Question-Thought-Action-Observation (ER-QTAO)'', which interprets user needs, retrieves railway service information from external sources, and generates satisfactory responses.

In the following subsections, we elaborate on each component of ER-QTAO and detail the railway services provided by LLM4Rail.

\subsection{Experience-Retrieval QTAO (ER-QTAO)}\label{sec:er_qtao}

The design of the ER-QTAO reasoning framework draws inspiration from two established paradigms: ReAct~\cite{react}, which interleaves reasoning and acting in LLM-based agents, and Reflexion~\cite{reflexion}, which accumulates long-term experiences through trajectory evaluation.
However, directly adapting these approaches to a railway-specific, multi-task, and dynamic scenario presents significant challenges. 
First, as the number and complexity of external tools grow, long-term memory occupies an excessive portion of the context window. 
Second, the framework lacks a mechanism to refresh stale experiences.

To address the limitations of ReAct-style tool orchestration, ER-QTAO proposes two key mechanisms: (i) Experience Storage and Retrieval (ESR), which accumulates structured failure-recovery cases from historical queries and retrieves relevant evidence to support current reasoning; (ii) Diagnostic Observation Generation (DOG), which transforms raw outputs into structured diagnostic reports to guide subsequent actions and facilitate experience summarization.

\subsubsection{(1) Iterative Reasoning Process.}

Consider a general scenario in which the LLM4Rail agent interacts with a passenger, who initiates the conversation by asking a railway-specific question (``Question'').
ER-QTAO first enables the agent to recognize the passenger's intent (``Thought''), then instructs it to select an appropriate API (``Action''), extract the required input parameters, and invoke the API to retrieve external evidence for railway service information (``Observation''). 
The retrieved information are subsequently fed back into the foundation LLM to generate the final response.

LLM4Rail evaluates whether the acquired information suffices to answer the user's question. 
If the retrieved information proves insufficient or unhelpful, LLM4Rail then triggers another round of function calls. 
This iterative process can be repeated multiple times until either a predefined maximum number of attempts is reached or sufficient information has been gathered.
Note that $\mathcal{T}^{t}_{i}$, $\mathcal{A}^{t}_{i}$, and $\mathcal{O}^{t}_{i}$ denote the contents of the ``Thought'', ``Action'', and ``Observation'' steps, respectively, at the $i$-th iteration of the $t$-th dialogue turn ($1\leq i \leq n$ and $1 \leq t \leq T$).

\subsubsection{(2) Prompt Construction.}\label{sec:prompt_construct}

At the $t$-th dialogue turn, the LLM is fed with a prompt that concatenates four components: the conversation history from the previous $t-1$ dialogue turns ($\mathcal{C}_{1:t-1}$), the base instruction ($\mathcal{B}$), the current user query ($Q^t$), and a set of experiences ($\mathcal{E}^{t}$). 
If the current query requires multiple iterations to resolve, the system sequentially appends $\mathcal{T}^{t}_{1}, \mathcal{A}^{t}_{1}, \mathcal{O}^{t}_{1} \dots \mathcal{T}^{t}_{n-1}, \mathcal{A}^{t}_{n-1}, \mathcal{O}^{t}_{n-1}$ from preceding iterations within the same dialogue turn to the prompt.

Formally, the prompt at the $n$-th iteration is:
\begin{equation}
    \begin{aligned}
        Prompt_{n}^{t} = [\,\mathcal{C}_{1:t-1}, \mathcal{B}, Q^t, \mathcal{E}^{t}, \mathcal{T}^{t}_{1}, \mathcal{A}^{t}_{1}, \mathcal{O}^{t}_{1}, \dots, \mathcal{T}^{t}_{n-1}, \mathcal{A}^{t}_{n-1}, \mathcal{O}^{t}_{n-1}\,],
    \end{aligned}
    \label{eq:er_qtao_prompt}
\end{equation}

The LLM then generates the next ``Thought'' $\mathcal{T}_{n}^{t}$ and ``Action'' $\mathcal{A}_{n}^{t}$:
\begin{equation}
    \mathcal{T}_{n}^{t}, \mathcal{A}_{n}^{t} \gets \text{LLM}(Prompt_{n}^{t} \mid \Theta).
    \label{eq:er_qtao_llm}
\end{equation}

If the LLM judges that enough information has been gathered, the action $\mathcal{A}_{n}^{t}$ yields the final answer $Ans^t$; otherwise, $\mathcal{A}_{n}^{t}$ specifies a new tool API to be invoked in the next iteration with corresponding input parameters:
\begin{equation}
    \mathcal{A}_{n}^{t} = 
    \begin{cases}
        Ans^{t}, \ \text{if the LLM determines that the query can be answered}, \\
        API_{n}^{t}\, (Input\_Parameter_{n}^{t}), \ \text{otherwise}.
    \end{cases}
    \label{eq:er_qtao_action}
\end{equation}

The raw results $Raw_{n}^{t}$ returned by the invoked external tools are subsequently processed by the DOG module to produce the diagnostic observation $\mathcal{O}_{n}^{t}$.
If the external tool API invocation fails and the failure is later resolved, the trajectory is summarized into a new experience and stored in the ESR module.

\subsubsection{(3) Experience Storage and Retrieval (ESR).}\label{sec:esr}

Summarized from historical interactions, the experiences capture how tools are used and fall into two categories: query-agnostic experience ($e_{agn}$) and query-related experience ($e_{rel}$). 
Query-agnostic experiences refer to the tool's general usage patterns, which are independent of the user's specific needs, while query-related experiences specify the appropriate actions for particular situations.
Mathematically, 
\begin{equation}
    e_{agn} = (tool,\, failure,\, diagnosis,\, strategy),
    \label{eq:def_experience_agn}
\end{equation}
\begin{equation}
    e_{rel} = (query,\, failure,\, diagnosis,\, strategy),
    \label{eq:def_experience_rel}
\end{equation}

\noindent where $tool$, $query$, $failure$, $diagnosis$, $strategy$ denote the tool name, the user query, a failed tool invocation example, the diagnostic analysis, and the LLM-summarized strategy, respectively.

Let $\mathcal{E}_{agn}$ and $\mathcal{E}_{rel}$ denote the sets of query-agnostic and query-related experiences collected from prior interactions, respectively. 
The experience set $\mathcal{E}^{t}$ at the $t$-th dialogue turn is the union of $\mathcal{E}_{agn}$ and $\mathcal{E}_{rel}^{t}$:
\begin{equation}
    \mathcal{E}^{t} = \mathcal{E}_{agn} \cup \mathcal{E}_{rel}^{t}.
\end{equation}

Note that $\mathcal{E}^{t}$ includes the entire query-agnostic experience set $\mathcal{E}_{agn}$ and the query-related experience subset $\mathcal{E}_{rel}^{t}$.
$\mathcal{E}_{rel}^{t}$ consists of the top-$k$ experiences in $\mathcal{E}_{rel}$ whose queries ($e_i.query$) are most similar to the $t$-th turn input question $Q^t$.
\begin{equation}
    \mathcal{E}_{rel}^{t} = \mathop{\arg\max}_{\mathcal{S} \subseteq \mathcal{E}_{rel}, |\mathcal{S}|=k} \sum_{e_i \in \mathcal{S}} \text{sim}\big(\text{embed}(Q^t), \text{embed}(e_i.{query})\big),
    \label{eq:esr_retrieve}
\end{equation}

\noindent where $\text{embed}(\cdot)$ denotes a text embedding function, and $\text{sim}(\cdot,\cdot)$ computes the cosine similarity between two embeddings. 
Then the experience set $\mathcal{E}^{t}$ is incorporated into the prompt (see Eq.\,\eqref{eq:er_qtao_prompt}) to guide the LLM to correctly invoke external tool APIs.

Upon loop termination, if any failure is diagnosed and subsequently resolved during the iterations, the failure-recovery pair is summarized into a new experience and stored in the sets:
\begin{equation}
    \mathcal{E}_{rel} \leftarrow \mathcal{E}_{rel} \cup \text{Summarize}_{rel}(Q^t, \mathcal{T}_1^t, \mathcal{A}_1^t, \mathcal{O}_1^t,..., \mathcal{T}_{n}^t, \mathcal{A}_{n}^t),
    \label{eq:esr_rel_store}
\end{equation}
\begin{equation}
    \mathcal{E}_{agn} \leftarrow \mathcal{E}_{agn} \cup \text{Summarize}_{agn}(\mathcal{E}_{agn}, Q^t, \mathcal{T}_1^t, \mathcal{A}_1^t, \mathcal{O}_1^t,..., \mathcal{T}_{n}^t, \mathcal{A}_{n}^t),
    \label{eq:esr_agn_store}
\end{equation}

\noindent where $\text{Summarize}(\cdot)$ denotes the experience summarization function that leverages the LLM to extract the experiences. 
Notably, the summarization function takes the existing $\mathcal{E}_{agn}$ as input to avoid extracting duplicate experiences.

In addition, any outdated experience found to mislead the current query will be detected and pruned from the experience set:
\begin{equation}
    \mathcal{E}_{rel} \leftarrow \mathcal{E}_{rel} \setminus \text{Outdated}_{rel}(\mathcal{E}_{rel}^t, Q^t, \mathcal{T}_1^t, \mathcal{A}_1^t, \mathcal{O}_1^t,..., \mathcal{T}_{n}^t, \mathcal{A}_{n}^t),
    \label{eq:esr_rel_outdate}
\end{equation}
\begin{equation}
    \mathcal{E}_{agn} \leftarrow \mathcal{E}_{agn} \setminus \text{Outdated}_{agn}(\mathcal{E}_{agn}, Q^t, \mathcal{T}_1^t, \mathcal{A}_1^t, \mathcal{O}_1^t,..., \mathcal{T}_{n}^t, \mathcal{A}_{n}^t).
    \label{eq:esr_agn_outdate}
\end{equation}

\subsubsection{(4) Diagnostic Observation Generation (DOG).}\label{sec:dog}
In naive ReAct, the observation returned to the LLM is the raw output of the external tool. 
In ER-QTAO, the raw output is first processed by DOG module to produce a structured diagnostic report. Specifically, after executing the action $\mathcal{A}_n^t$, the external tool returns a raw response $Raw_n^t$. 
Then, the DOG module transforms $Raw_n^t$ into a structured observation $\mathcal{O}_n^t$:
\begin{equation}
    \mathcal{O}_{n}^{t} = \text{DOG}(Raw_{n}^{t}, \mathcal{A}_{n}^{t}) = 
    \begin{cases}
        \{\text{Status}: \text{Success},\,\text{Result}: Raw_{n}^{t}\},  \text{if } \text{Val}(Raw_{n}^{t}) = \text{True}, \\
        \{\text{Status}: \text{Fail},\,\text{Cause}: c,\, \text{Hint}: h\}, \, \text{otherwise},
    \end{cases}
    \label{eq:dog}
\end{equation}

\noindent where $\text{Val}(\cdot)$ is a validation function that checks whether the raw output indicates successful execution (e.g., non-empty result set). In case of failure, DOG performs sequential diagnostic checks to identify the root cause $c$ (e.g., \textsc{Station\_Invalid}, \textsc{Route\_Invalid}), and generates a human-readable hint containing a suggested correction $h$ (e.g., ``Reachable destinations from Shanghai Station: `Beijing Station', `Guangzhou Station', etc.'') for the LLM.

\subsection{Railway-Specific Services Offered by LLM4Rail}

\subsubsection{(1) Ticket Inquiry.} 

Ticket inquiry is among the most basic functions of online railway travel services. 
Passengers typically enter key parameters, such as the departure station, travel date, and destination. 
These inputs are then used to search the database for the best matching train service.
Once LLM4Rail recognizes the user's intent to inquire about ticket information, it triggers the ``Ticket Inquiry'' module at the ``Action'' step.
Furthermore, LLM4Rail supports fuzzy search for ticket information.
For example, when a passenger checks train ticket availability from ``Chongqing'' to ``Chengdu'' for tomorrow afternoon via the ER-QTAO framework, if no exact match is found between ``Chongqing Station'' and ``Chengdu Station'', the system automatically suggests alternatives.
This fallback mechanism is triggered by an error message from DOG module at the ``Observation'' stage, signaling that no direct service exists between the specified stations in the first iteration.
In the next iteration, LLM4Rail will query the closest alternative route (e.g., from ``Chongqing North Station'' to ``Chengdu East Station''). 
In the subsequent round, three different train options are displayed for the passenger to choose from.

\subsubsection{(2) Weather Inquiry.}
Weather conditions are very important for rail travel. 
Before starting trips, passengers often check the real-time weather forecast for their destination so that they can decide whether to bring an umbrella or warm clothing.
However, one significant challenge in the offline deployment of LLMs is the time efficiency issue, particularly in real-time applications.
Consequently, when passengers inquire about weather conditions directly from LLMs, they either fabricate information or fail to provide relevant details.
To address this limitation, LLM4Rail integrates external tools to offer weather service.
Once the LLM4Rail agent detects that a passenger intends to check the weather forecast, it triggers the action planning process to call the weather module to obtain relevant weather information. 
The agent leverages the LLM's strong semantic understanding capabilities to perform slot filling, extracting key details such as the time and location for weather inquiries.
For instance, if a passenger asks, ``What's the weather in Beijing tomorrow?'', the LLM4Rail agent extracts the keywords `tomorrow' and `Beijing', groups them into the following format: $\{time\!: \langle tomorrow \rangle, \ place\!: \langle Beijing \rangle \}$, and passes this information to the weather module.
The weather module then calls an \href{https://lbs.amap.com/api/webservice/guide/api/weatherinfo}{API} provided by the Amap Open Platform to retrieve the weather information.

\subsubsection{(3) Food \& Drink Recommendation.} 
Online food ordering and delivery services benefit the railway sector by providing personalized onboard services while generating additional revenue.
Passengers place orders online before departure, and restaurants prepare and deliver the food to the train upon order receipt, enabling passengers to enjoy their meals after boarding.

\begin{figure*}[t]
    \hspace{-1cm}
    \includegraphics[width=14cm]{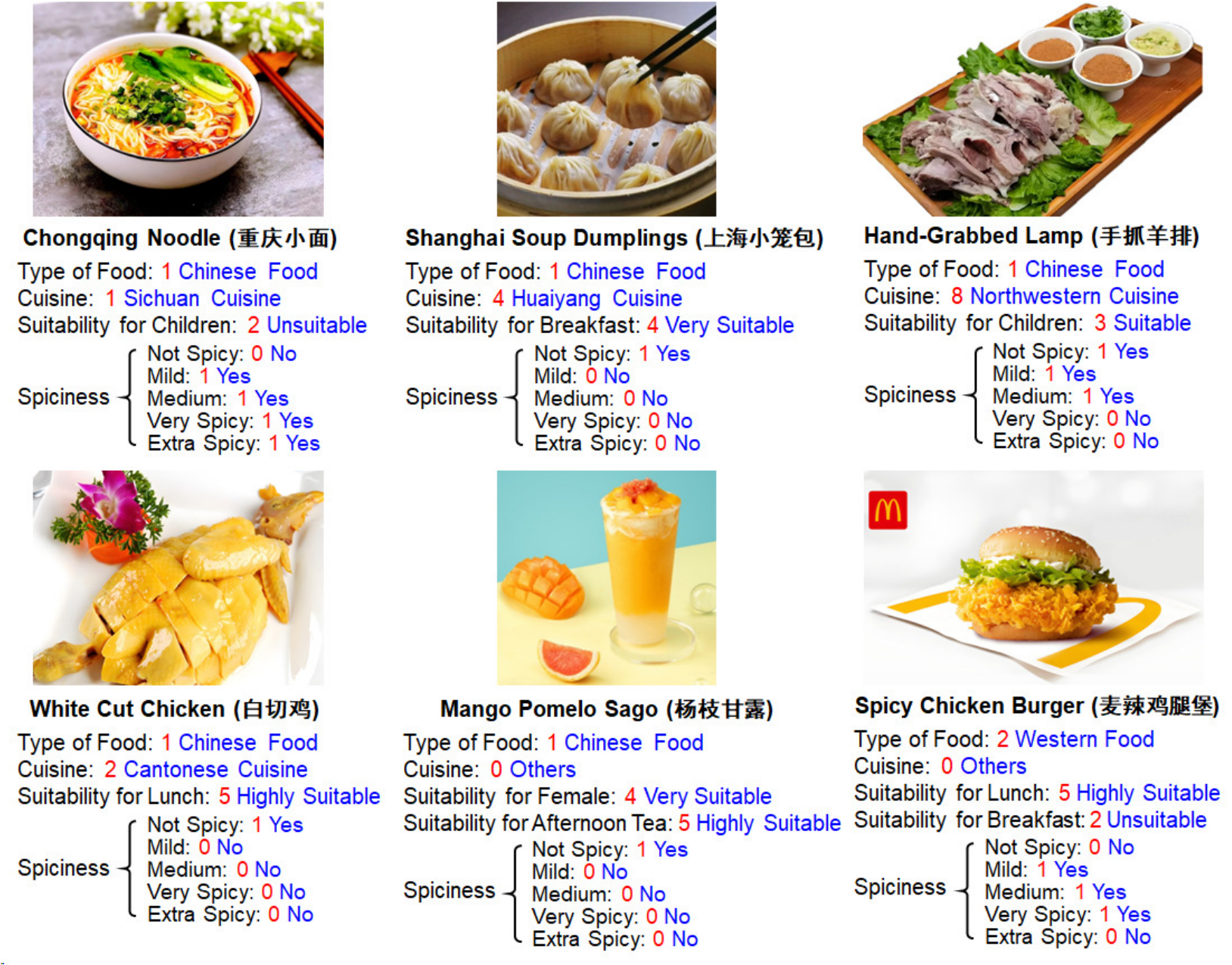}
    \caption{An illustration of the Chinese Railway Food and Drink (CRFD) takeout dataset constructed by us.} 
    \label{illustration_of_crfd-25}
\end{figure*}

\noindent \textbf{$\bullet$ The Chinese Railway Food and Drink dataset (CRFD).}
In this section, we first introduce CRFD, a novel dataset designed to advance food \& drink recommendation for railway travelers. 
It encompasses representative dishes from 33 major railway hub cities, with over 200 signature dishes manually curated and labeled per city across various restaurants. 
Data were collected from leading Chinese online food delivery platforms.

The dish selection, as depicted in Fig.~\ref{illustration_of_crfd-25}, adhered to two principles: (i) balancing regional specialties with widely consumed fast food, and (ii) catering to diverse age groups. For example, Shanghai's entries include both Shanghai Soup Dumplings (traditional local cuisine) and McDonald's (global fast food), alongside hamburgers (child-friendly) and Mango Pomelo Sago (popular among younger demographics).
Each CRFD entry comprises an illustrative image and structured feature annotations. 
Key attributes include food type, cuisine, meal suitability, child-friendliness, and spiciness. Categorical features are encoded as integer labels\,--\,e.g., food type: 1 (Chinese), 2 (Western); cuisine: 1 (Sichuan), 2 (Cantonese), 3 (Shandong). 
Multi-select features adopt a multi-hot encoding scheme. 
For example, Chongqing Noodles' spiciness feature marks available levels (mild through extra spicy) as 1 and not spicy as 0.

For more detailed information about CRFD, please refer to the Github repository at \href{https://www.github.com/CQUPTAI/CRFD-25}{https://www.github.com/CQUPTAI/CRFD-25}.

\begin{figure*}[t]
    \hspace{-3.2cm}
    \includegraphics[width=18.5cm]{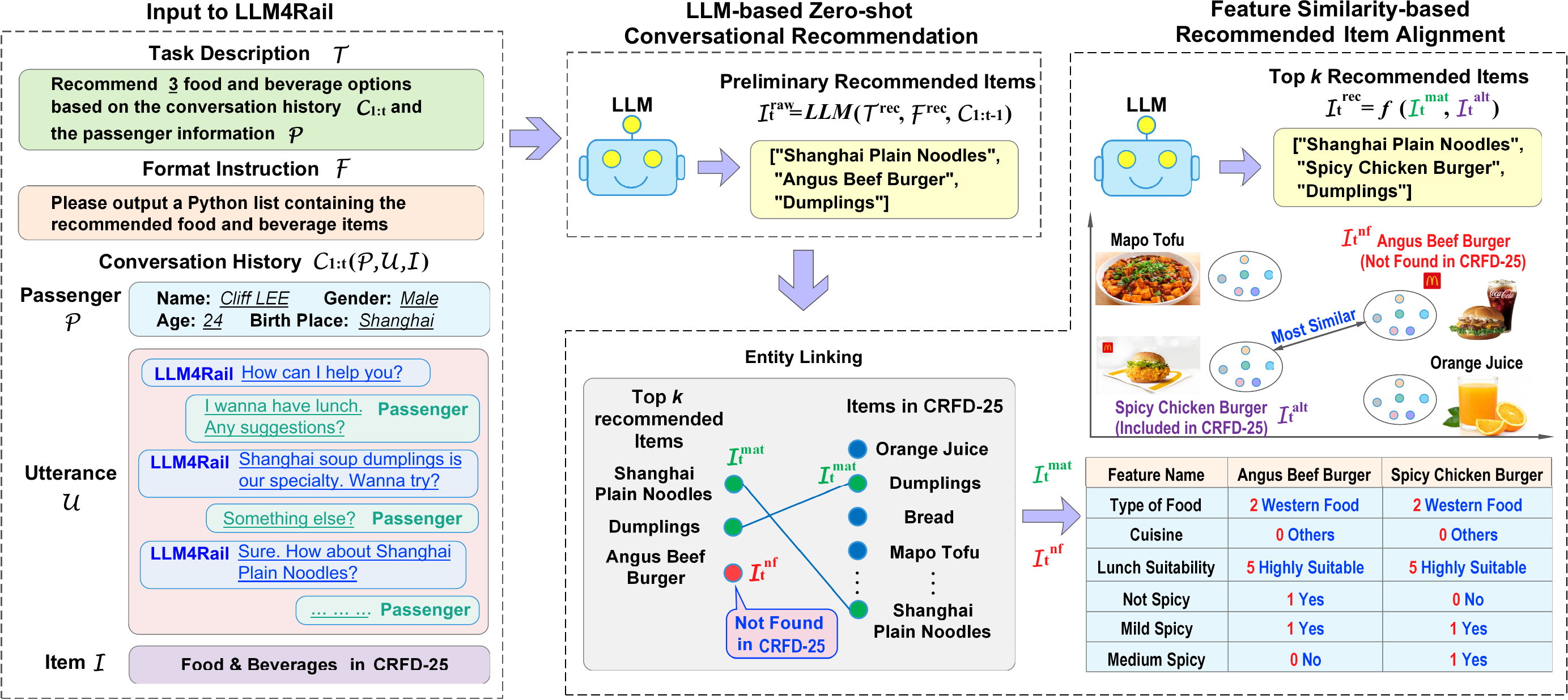}
    \caption{An illustration of the LLM-based conversational food \& drink recommendations using feature similarity-based item alignment.} 
    \label{fig:illustration_of_recommendation}
\end{figure*}

\noindent \textbf{$\bullet$ LLM-based Conversational Recommendation.}
To address the food and beverage recommendation issue within the context of online railway meal ordering, we propose a novel LLM-based conversational recommendation algorithm that integrates traditional content-based approaches~\cite{unbert,DSSM,KGCN} with LLM-based  conversational methods~\cite{zeroshot-crs,rta,crag,memocrs,usb-rec,grag-crs,rag-crs}.

\textbf{LLM-based Conversational Recommender System:} Given a passenger set $\mathcal{P}$, an utterance set $\mathcal{U}$ and an food and beverage item set $\mathcal{I}$ in CRFD, a passenger $p_t \in \mathcal{P}$ generates an utterance $u_t \in \mathcal{U}$ during the $t$-th round of conversation, where $t = 1, 2, \dots, T$.
The utterance $u_t$ may contains a set of items $\mathcal{I}_t \in \mathcal{I}$, where $\mathcal{I}_t$ can be an empty set if no items are mentioned in $u_t$.
The complete conversation is then denoted as $\mathcal{C}_{1:T} = (p_t, u_t, \mathcal{I}_t)_{t=1}^T$.
Building upon the approach in~\cite{zeroshot-crs}, we utilize LLMs as a zero-shot conversational recommender system.
At the $t$-th turn of the conversation, the LLM-based conversational recommender inputs the dialogue history $\mathcal{C}_{1:t-1} = (p_t, u_t, \mathcal{I}_t)$ into the LLM to generate top-$k$ of preliminary recommended items $\mathcal{I}_{t}^{raw}$.
Alongside the conversation history, the task description $\mathcal{T}^{rec}$ and the format instruction $\mathcal{F}^{rec}$ are fed into the LLM as additional inputs. 
This can be expressed as follows:
\begin{equation}
    \mathcal{I}_{t}^{raw}=LLM(\mathcal{T}^{rec},\mathcal{F}^{rec},\mathcal{C}_{1:t-1}),
    \label{eq_LLM_recommend}
\end{equation}

\noindent where $\mathcal{T}^{rec}$ prompts the LLM to recommend food \& beverages of potential interest to the passengers based on the dialogue context and their profile (e.g., gender, age, and home city), whereas $\mathcal{F}^{rec}$ enforces a predefined output format.

\indent \textbf{Entity Linking:} The conventional zero-shot conversational recommendation achieves promising performance in movie recommendation~\cite{zeroshot-crs}, leveraging LLMs' prior knowledge and linguistic capabilities. 
However, it suffers from a critical limitation in our scenario: LLM-generated items may be sold out or nonexistent (i.e., $\mathcal{I}_{t}^{raw} \nsubseteq \mathcal{I}$). 
To address this, we first map the raw recommendation set $\mathcal{I}_{t}^{raw}$ to a subset of $\mathcal{I}$:
\begin{equation}
    \mathcal{I}_{t}^{mat}= \Phi(\mathcal{I}_{t}^{raw}, \mathcal{I})
    \label{eq_entity_linking}
\end{equation}

\noindent where the mapping function $\Phi$ is based on string matching or semantic matching and finds the nearest neighbor for each preliminary item in $\mathcal{I}^{raw}_{t}$ to get the matched items $\mathcal{I}_{t}^{mat}$ existing in the dataset.
Furthermore, $\mathcal{I}_{t}^{mat}$ is input into the LLM to identify the matched items that are irrelevant to the corresponding preliminary items:
\begin{equation}
    \mathcal{I}^{irr}_{t}, \mathcal{I}_{t}^{nf} = LLM(\mathcal{I}_{t}^{mat}, \mathcal{I}_{t}^{raw}, \mathcal{T}^{ide}, \mathcal{F}^{ide})
    \label{eq_remove_irrelevant}
\end{equation}

\noindent where $\mathcal{I}_{t}^{irr}$ is the irrelevant items in $\mathcal{I}_{t}^{mat}$ and $\mathcal{I}_{t}^{nf}$ are their counterparts in $\mathcal{I}_{t}^{raw}$, while $\mathcal{T}^{ide}$ and $\mathcal{I}^{ide}$ are the task description and format instruction, respectively. $\mathcal{I}^{irr}_{t}$ will be removed from the matched item set $\mathcal{I}_{t}^{mat}$.

\indent \textbf{Fallback Strategy:} After identifying the items that are not found in the dataset $\mathcal{I}_{t}^{nf}$, we leverage their content information to retrieve alternatives for their irrelevant matched items.
The LLM is leveraged to generate the attributes aligned with the CRFD dataset format for each of them:
\begin{equation}
     \mathcal{A}_{t} = LLM(\mathcal{I}_{t}^{nf}, \mathcal{T}^{ann}, \mathcal{F}^{ann}),
    \label{eq_fallback_annotate}
\end{equation}

\noindent where $\mathcal{A}_{t}$ is the attributes of each unmatched item annotated by the LLM, whereas $\mathcal{T}^{ann}$ and $\mathcal{F}^{ann}$ are the annotation task description and output format requirement.
Then the attributes are utilized to retrieve the most similar items from the item set $\mathcal{I}$:
\begin{equation}
    \mathcal{I}_{t}^{alt,k} = \arg\min_{I}sim(\mathcal{A}_{t}^{k}, \mathcal{A}_{I}),
    \label{eq_fallback_retrieve}
\end{equation}

\noindent where $\mathcal{A}_{t}^{k}$ is the generated attributes of the $k$-th unmatched item, $\mathcal{A}_{I}$ is the attributes of item $I \in \mathcal{I}$, and $\mathcal{I}_{t}^{alt,k}$ is the retrieved alternative item for the k-th unmatched item. 
Then we can obtain the final recommended items:
\begin{equation}
    \mathcal{I}_{t}^{rec} = \mathcal{I}^{mat}_{t}\cup(\bigcup_{k}\mathcal{I}_{t}^{alt,k})
\end{equation}

Fig.~\ref{fig:illustration_of_recommendation} illustrates the recommendation process, where ``Angus Beef Burger'' cannot be found in CRFD and retrieve ``Spicy Chicken Burger'' as an alternative. 
In this way, each candidate item recommended by the LLM is aligned with an available item in the dataset, ensuring the effectiveness of food and beverage recommendations by LLM4Rail.

\subsubsection{(4) Casual Conversation.}
In addition to the railway travel-related services mentioned above, LLM4Rail also equips a chitchat module that allows passengers to engage in casual conversations with LLM.
Instead of building our own dialog system, the chitchat module depends entirely on the LLM's abilities and extensive knowledge to facilitate casual conversations on a wide range of topics, including movies, sports, travel, and more.

\section{Experiment}

\subsection{Experimental Setup}

The system is evaluated on three key aspects: ticket inquiry, weather inquiry, and food \& beverage recommendation.
We leverage three publicly available foundation models: \verb|Qwen3-235b-a22b|, \verb|DeepSeek-v3.1| and \verb|ChatGPT-4o|. 
For these models, all decoding hyper-parameters are set to their default values as recommended in the official APIs.

\begin{table*}[t]
\caption{Performance comparison of various reasoning frameworks and foundation models on Ticket Inquiry and Weather Inquiry tasks in terms of information retrieval accuracy (Acc) and average number of reasoning iterations ($\#\overline{\mbox{Iter}}$). }
\hspace{-1.0cm}
{
{
\begin{tabular}
{
  c 
  !{\vrule width 1.5pt}l  
  !{\vrule width 1.5pt}c  
  |c  
  !{\vrule width 1.5pt}c  
  |c  
  !{\vrule width 1.5pt}c  
  |c  
  !{\vrule width 1.5pt}c  
  |c  
  !{\vrule width 1.5pt}c  
  |c
}
\toprule
\renewcommand{\arraystretch}{2}
\multirow{2}{*}{\raisebox{-1.5ex}{\hspace{-0.1cm}\ \scalebox{1}[1]{\textbf{Module}} }}
 &  \multirow{2}{*}{\raisebox{-2ex}{\hspace{0.05cm}\textbf{\makecell{Foundation\\LLM}}}}
& \multicolumn{2}{c!{\vrule width 1.5pt}}{\scalebox{0.9}[1]{\textbf{LLM-Only}} } & \multicolumn{2}{c!{\vrule width 1.5pt}}{\textbf{CoT}\cite{chain_of_thought}} & \multicolumn{2}{c!{\vrule width 1.5pt}}{\textbf{ReAct}\cite{react}} & \multicolumn{2}{c!{\vrule width 1.5pt}}{\scalebox{0.9}[1]{\textbf{Reflexion}\cite{reflexion}}} & \multicolumn{2}{c}{\textbf{LLM4Rail}} \\
\cmidrule(lr){3-4} \cmidrule(lr){5-6} \cmidrule(lr){7-8} \cmidrule(lr){9-10} \cmidrule(lr){11-12}
&    & \scalebox{0.8}[1]{\textbf{Acc\,(\%)}} & \textbf{\,\scalebox{0.6}[1]{\#}\scalebox{0.8}[1.1]{$\overline{\text{Iter}}$}\,} & \scalebox{0.8}[1]{\textbf{Acc\,(\%)}} & \textbf{\,\scalebox{0.6}[1]{\#}\scalebox{0.8}[1.1]{$\overline{\text{Iter}}$}\,} & \scalebox{0.8}[1]{\textbf{Acc\,(\%)}} & \textbf{\,\scalebox{0.6}[1]{\#}\scalebox{0.8}[1.1]{$\overline{\text{Iter}}$}\,} & \scalebox{0.8}[1]{\textbf{Acc\,(\%)}} & \textbf{\,\scalebox{0.6}[1]{\#}\scalebox{0.8}[1.1]{$\overline{\text{Iter}}$}\,} & \scalebox{0.8}[1]{\textbf{Acc\,(\%)}} & \textbf{\,\scalebox{0.6}[1]{\#}\scalebox{0.8}[1.1]{$\overline{\text{Iter}}$}\,} \\
\midrule
\multirow{3}{*}{\textbf{\makecell {\scalebox{0.95}[1]{Ticket}\\\scalebox{0.9}[1]{Inquiry}}  }}
 & \scalebox{0.88}[1]{\textbf{Qwen3-235B}} & 23.6\% & 1.00 & 48.4\% & 1.00 & 21.8\% & 2.30 & 19.8\% & 1.00 & 24.8\% & 2.92 \\
 & \scalebox{0.85}[1]{\textbf{DeepSeek-v3.1}} & 45.8\% & 1.00 & 64.2\% & 1.00 & 52.0\% & 3.45 & 28.6\% & 1.00 & 60.4\% & 3.50 \\
 & \scalebox{0.85}[1]{\textbf{ChatGPT-4o}} & 38.5\% & 1.00 & 52.3\% & 1.00 & 42.7\% & 2.26 & 25.0\% & 1.00 & 51.6\% & 2.90 \\
\midrule
\multirow{3}{*}{\textbf{\makecell{ \scalebox{0.9}[1]{Weather} \\ \scalebox{0.9}[1]{Inquiry}}  }}
 & \scalebox{0.88}[1]{\textbf{Qwen3-235B}} & 36.0\% & 1.00 & 52.3\% & 1.00 & 60.2\% & 2.64 & 45.8\% & 1.00 & 63.1\% & 2.98 \\
 & \scalebox{0.85}[1]{\textbf{DeepSeek-v3.1}} & 37.0\% & 1.00 & 54.6\% & 1.00 & 71.7\% & 3.00 & 48.3\% & 1.00 & 74.2\% & 3.20 \\
 & \scalebox{0.88}[1]{\textbf{ChatGPT-4o}} & 40.8\% & 1.00 & 61.3\% & 1.00 & 78.2\% & 2.98 & 54.6\% & 1.00 & 81.9\% & 3.00 \\
\bottomrule
\end{tabular}%
}
}
\label{tab:model_comparison}
\end{table*}

\begin{table*}[t]
    \hspace*{-1.3cm} 
    \begin{minipage}[t]{7.4cm}
        \centering
        \caption{Ablation study: Impact of ESR and DOG on information retrieval accuracy for ticket inquiry.}
        \label{tab:ablation}
        \renewcommand{\arraystretch}{1.3}
        \begin{tabular}{p{1.9cm}<{\centering} p{1.4cm}<{\centering} p{1.45cm}<{\centering} p{1.75cm}<{\centering}}
            \toprule
            \textbf{\makecell{Reduced\\Model}}
            & \scalebox{0.9}[1]{\textbf{\makecell{Qwen3\\-235B}}} 
            & \scalebox{0.9}[1]{\textbf{\makecell{DeepSeek\\-v3.1}}} 
            & \scalebox{0.9}[1]{\textbf{\makecell{ChatGPT\\-4o}}} \\
            \midrule
            \scalebox{0.9}[1]{\textbf{ER-QTAO}} & \textbf{24.8\%} & \textbf{60.4\%} & \textbf{51.6\%} \\
            \scalebox{0.9}[1]{\textbf{w/o ESR}} & 22.2\% & 54.7\% & 45.6\% \\
            \scalebox{0.9}[1]{\textbf{w/o DOG}} & 23.4\% & 58.0\% & 49.5\% \\
            \scalebox{0.9}[1]{\textbf{w/o Both}} & 21.8\% & 52.0\% & 42.7\% \\
            \bottomrule
        \end{tabular}
    \end{minipage}
    \hspace{0.2cm}
    \begin{minipage}[t]{7cm}
        \centering
        \caption{Failure reason analysis for the ticket inquiry and weather inquiry tasks using GPT-4o.}
        \label{tab:failure_reason}
            \begin{tabular}{p{1.5cm}<{\centering} p{1.7cm}<{\centering} p{1.6cm}<{\centering} p{1.5cm}<{\centering}}
            \toprule
                \multirow{3}{*}{\makecell{ \scalebox{0.95}[1]{\textbf{Failure}} \\ \scalebox{0.95}[1]{\textbf{Cause}}  }} & \scalebox{1}[1]{\textbf{Ticket}} & \scalebox{1}[1]{\textbf{Ticket}} & \multirow{3}{*}{\makecell{ \scalebox{0.95}[1]{\textbf{Weather}} \\ \scalebox{1}[1]{\textbf{(\%)}}  }} \\
                & \scalebox{0.9}[1]{\textbf{w/o}} \scalebox{0.85}[1]{\textbf{DOG}} & \scalebox{0.9}[1]{\textbf{w/ DOG}} &  \\
                & \scalebox{0.95}[1]{\textbf{(\%)}} & \scalebox{0.95}[1]{\textbf{(\%)}} &  \\
            \midrule
                \textbf{`Station'} & 13.18\% & 27.36\% & - \\
                \textbf{`City'} & - & - & \textbf{61.90\%} \\
                \textbf{`Date'} & \textbf{55.43\%} & \textbf{45.28\%} & 38.10\%  \\
                \textbf{`Time'} & 31.39\% & 27.36\% & - \\
            \bottomrule
        \end{tabular}
    \end{minipage}
\end{table*}

\subsection{Experimental Results}

\subsubsection{Results for Ticket Inquiry and Weather Inquiry:}

Table~\ref{tab:model_comparison} presents a performance comparison of ticket inquiry and weather inquiry across various foundation models.
As the table shows, across all baselines, LLM4Rail with DeepSeek achieves higher overall performance than the version using Qwen3 and ChatGPT for ticket inquiry.
Compared to ReAct, LLM4Rail achieves performance gains of 3.0\%, 8.4\% and 8.9\% for Qwen3, DeepSeek, and ChatGPT, respectively.
This improvement is attributed to the utilization of stored experiences and the structured diagnostic analysis.
In addition, LLM4Rail with EA-QTAO reasoning framework requires 0.43 more reasoning loops on aveage than ReAct.
Regarding weather information retrieval, across all baselines, the ChatGPT-based instantiation outperforms those using Qwen3 and DeepSeek.

Table~\ref{tab:ablation} presents the ablation results, comparing LLM4Rail with variants where ESR, DOG, or both are removed.
The complete version of LLM4Rail surpasses its ablated variants by a large margin, validating the effectiveness of the ESR and DOG modules.
Compared to ablating both modules, retaining only DOG yields higher accuracy, suggesting that DOG alone can improve the LLM's ability to invoke tools by explicitly prompting how to repair failed API calls.

Table~\ref{tab:failure_reason} summarizes the reasons why LLM4Rail fails in ticketing and weather.
In the context of ticket inquiries, we observe that over 70\% of failed query attempts are due to the inability to successfully resolve `Date' or `Time' parameters.
Our preliminary analysis suggests that current foundational models demonstrates limited temporal reasoning ability.
This issue can be further addressed by incorporating an external ``\emph{datetime}'' module.

\begin{figure*}[t]
    \hspace{-3.2cm}
        \includegraphics[width=18.5cm]{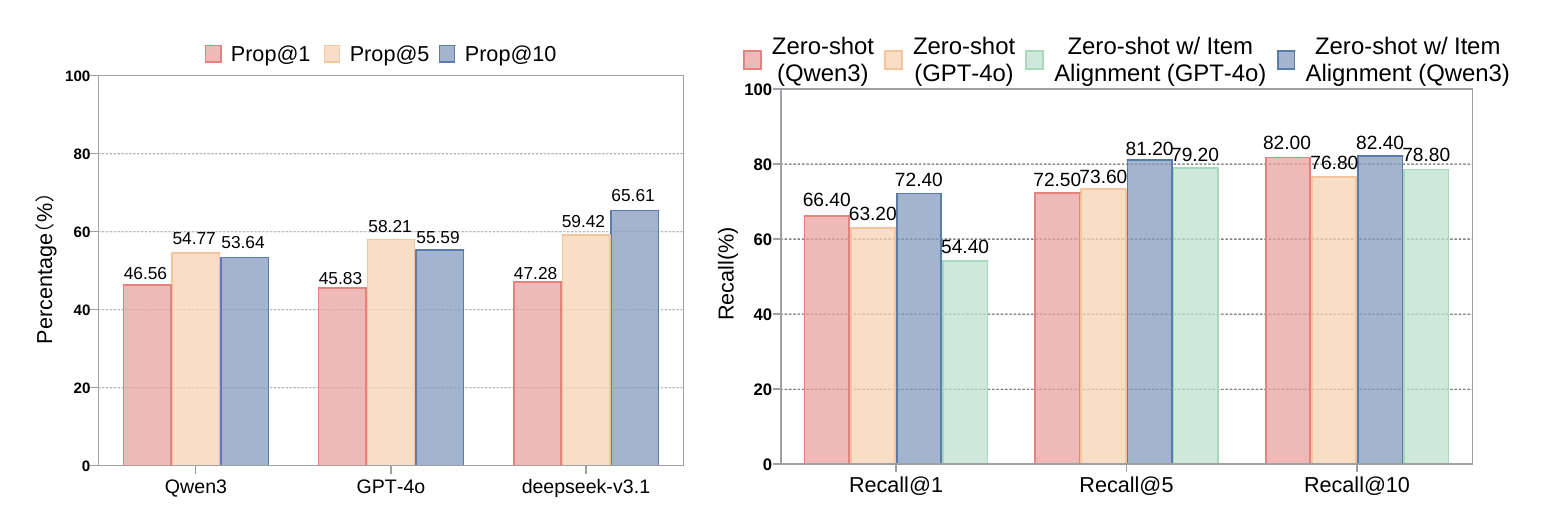}
        \vspace{-0.3cm}
        \caption{The proportion of the top-$k$ food and drink items $Prop@k \ (k=1,5,10)$ recommended by different foundation models that are present in the CRFD dataset (left), and the performance of food and drink recommendation by different recommendation methods in terms of $Recall@k \ (k=1,5,10)$ (right).} 
        \vspace{-0.1cm}
        \label{fig:bars_of_recommendation}
\end{figure*}

\begin{figure*}[t]
    \hspace{-3cm}
    \includegraphics[width=18.5cm]{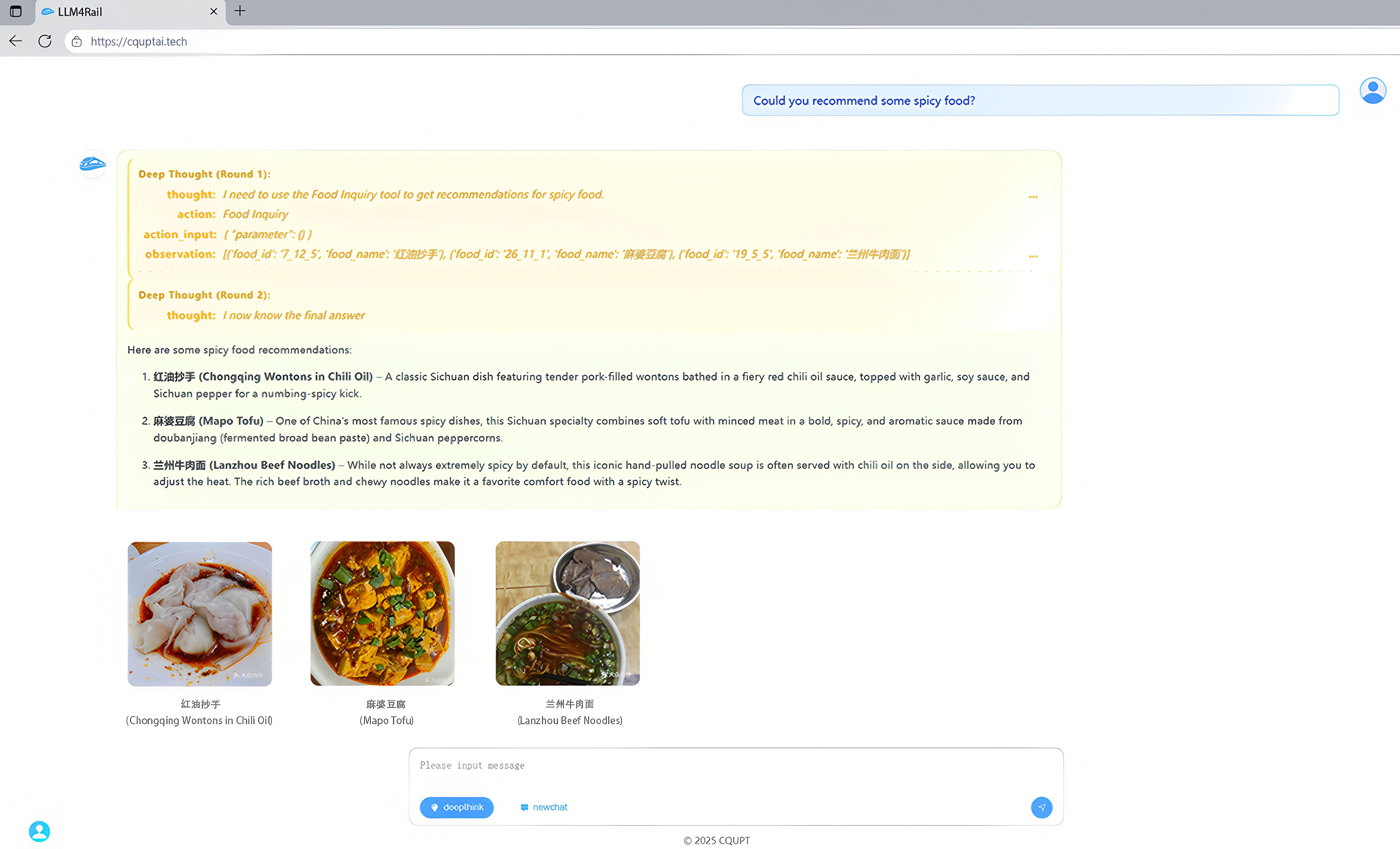}
    \caption{LLM4Rail system demonstration: Food \& Drink Recommendation.}
    \label{fig:casestudy_of_food_ordering}
\end{figure*}

\subsubsection{Results for Food \& Drink Recommendation:}

Fig.~\ref{fig:bars_of_recommendation} shows $Prop@k$, i.e., the proportion of top-$k$ $(k=1,5,10)$ food \& drink recommendations that are present in the CRFD dataset across different foundation models. 
It is notable that less than 60\% recommended food \& drink items appear in the CRFD dataset for the three advanced foundation models and all $k$ values. 
Although expanding the corpus may cover more recommendations, in practice, only a limited number of items can be recommended due to spatial and temporal constraints.
This finding underscores the necessity of aligning recommendations with item corpus.

To validate our feature alignment approach, we evaluate the recommendation performance using $Recall@k \ (k=1, 5, 10)$, consistent with prior works~\cite{zeroshot-crs,unleashing-crs}.
Due to the scarcity of dialogues in the field of food \& drink recommendation, we employ a user simulator~\cite{rethink-crs,usb-rec} to interact with the conversational recommenders. 
Fig.~\ref{fig:bars_of_recommendation} validates the effectiveness of our feature-alignment-based recommendation approach, yielding performance improvements of 6.0\%, 8.7\%, and 0.4\% at $k = 1, 5, 10$, respectively, compared to the variant without item alignment using Qwen3, and 5.6\% and 2.0\% at $k = 5$  and $k = 10$ for GPT-4o, respectively.
Notably, Qwen3 outperforms GPT-4o across all settings, which may be attributed to its special optimization for Chinese long-tail knowledge.

\subsubsection{LLM4Rail System Demonstration:}
A screenshot of the LLM4Rail interface is presented in Fig.~\ref{fig:casestudy_of_food_ordering}.
It gives an illustrative example of the conversation between a passenger and the LLM4Rail agent.
The passenger inquires about local specialties and can then enjoy the selected dishes after boarding the train. 
LLM4Rail subsequently recommends three signature dishes that suit the passenger's taste.

\section{Conclusion}
In this work, we present LLM4Rail, a railway passenger service system that features the novel ER-QTAO reasoning framework, the curated CRFD dataset, and the zero-shot conversational recommender system.
It is a valuable attempt to apply LLM agents to the railway sector.

\section{Acknowledgement}
This work was supported by the Ph.D. Scientific Research Foundation of Chongqing University of Posts and Telecommunications (Grant No. E012A2022026), the National Natural Science Foundation of China (Grant No. 62576055), the Natural Science Foundation of Chongqing (Grant No. CSTB2025YITP-QCRCX0010), and Lingnan University (Grant No. SDS24A17).

%
%
%
%
\bibliographystyle{splncs04}
\bibliography{myrefs}
%




\end{document}